\documentclass{article}

\PassOptionsToPackage{numbers, compress}{natbib}

\usepackage[main, final]{main}

\usepackage[utf8]{inputenc} % allow utf-8 input
\usepackage[T1]{fontenc}    % use 8-bit T1 fonts
\usepackage{hyperref}       % hyperlinks
\usepackage{url}            % simple URL typesetting
\usepackage{booktabs}       % professional-quality tables
\usepackage{amsfonts}       % blackboard math symbols
\usepackage{nicefrac}       % compact symbols for 1/2, etc.
\usepackage{microtype}      % microtypography
\usepackage{xcolor}         % colors
\usepackage{graphicx}
\usepackage{subcaption}  % in preamble
\usepackage{amsmath}
\usepackage{wrapfig}
\usepackage{float}
\usepackage{mdframed}
\hypersetup{hidelinks}

\title{Sparse Layers are Critical to Scaling \\ Looped Language Models}

\author{%
  Ryan Lee$^{1}$\thanks{Corresponding author. \texttt{ryantlee@usc.edu}}
  \qquad
  Jacob Biloki$^{2}$
  \qquad
  Edward J. Hu$^{3}$
  \qquad
  Jonathan May$^{1}$
  \\[0.5em]
  $^{1}$USC Information Sciences Institute \\
  $^{2}$Netflix \\
  $^{3}$Independent Researcher
}

\begin{document}

\maketitle
\vskip -0.5em

\begin{abstract}
Looped language models repeat a set of transformer layers through depth, reducing memory costs and providing natural early-exit points at loop boundaries. However, looped models do not scale as favorably as standard transformers with unique layers. We compare standard and Mixture-of-Experts (MoE) transformers, with and without looping, and find two main results. First, we find Looped-MoE models scale better than the standard baseline while dense looped models do not. We trace this to routing divergence between loops: in Looped-MoE models, different experts are activated on each pass through the same shared layers, recovering expressivity without additional parameters. Our second finding is that looped models have better compute-quality trade-offs with early exits than standard models. Because each loop ends with the same layers that produce the final output, loop boundaries are superior exit points, as confirmed by earlier output convergence at these points. In sum, we provide a clear direction for scaling looped models: a Looped-MoE model with early exits can not only beat standard transformers at scale, but also enable significant memory and inference savings with minimal degradation in quality. Code: \href{https://github.com/ml-ryanlee/looped-moe-scaling}{\texttt{github.com/ml-ryanlee/looped-moe-scaling}}.
\end{abstract}

\section{Introduction}
Language models (LMs) are expensive to store and slow during inference due to their size. This stems from a fundamental architectural limitation: LMs can consist of billions of parameters arranged as a sequential stack of computational layers, yet each parameter is only used once during an inference step. Re-using parameters through compute depth, as explored in universal transformers~\citep{dehghani_universal_2019}, LoopLMs~\citep{zhu_scaling_2025}, and recurrent models~\citep{geiping_scaling_2025}, is an appealing alternative, however such architectures have been found to under-perform baselines when compared on equal training compute~\citep{kaplan_scaling_2020}. 

The most common pattern in parameter re-use is to loop transformer layers through depth. We hypothesize the expressiveness bottleneck for looped LMs is the dense feed-forward network (FFN), which applies the same operation to every token, an invariance that compounds when looped. Sparse Mixture-of-Experts (MoE) \citep{shazeer_outrageously_2017} layers offer a natural remedy by routing tokens to specialized experts, introducing input-dependent computation that may recover expressiveness lost to weight tying. MoE has been combined with looping before \citep{csordas_moeut_nodate}, but alongside other architectural changes. 

In this work, we isolate the effect of sparse experts on looped scaling laws with a controlled study. We find that \textbf{(i) sparse experts resolve the scaling deficit} of looped dense layers and achieve the best downstream accuracy. Although looping reduces storage, it does not on its own address the inference cost of deep transformers. Thus we explore early exiting~\citep{teerapittayanon_branchynet_2016, xin_deebert_2020, schuster_confident_2022}, allowing tokens to exit before full depth, and find that \textbf{(ii) looped models have better compute-quality trade-offs} than standard transformers using training-free early exit.

We identify two key properties of sparse looped models which enable superior scaling and early exit trade-offs: \textbf{(i) Loop-specific expert routing.} In Looped-MoE models, the router selects a different set of experts for the same physical layer on each loop pass, enabling shared layers to implement distinct computations across iterations. \textbf{(ii) Earlier convergence at loop boundaries.} In looped models, more tokens reach near-final predictions after the first loop layers, meaning early exits are more accurate than in non-looped models.
\begin{mdframed}
\textbf{Practical Recommendation.} Replace dense FFNs with top-k MoE layers in looped LMs. The resulting Looped-MoE architecture scales better than a standard dense LM, stores fewer weights for the same accuracy, and enables more compute savings via early exits at loop boundaries.
\end{mdframed}

\begin{figure*}[t]
    \centering
    \begin{minipage}[t]{0.35\linewidth}
        \centering
        \includegraphics[height=2.75in,width=\linewidth,keepaspectratio]{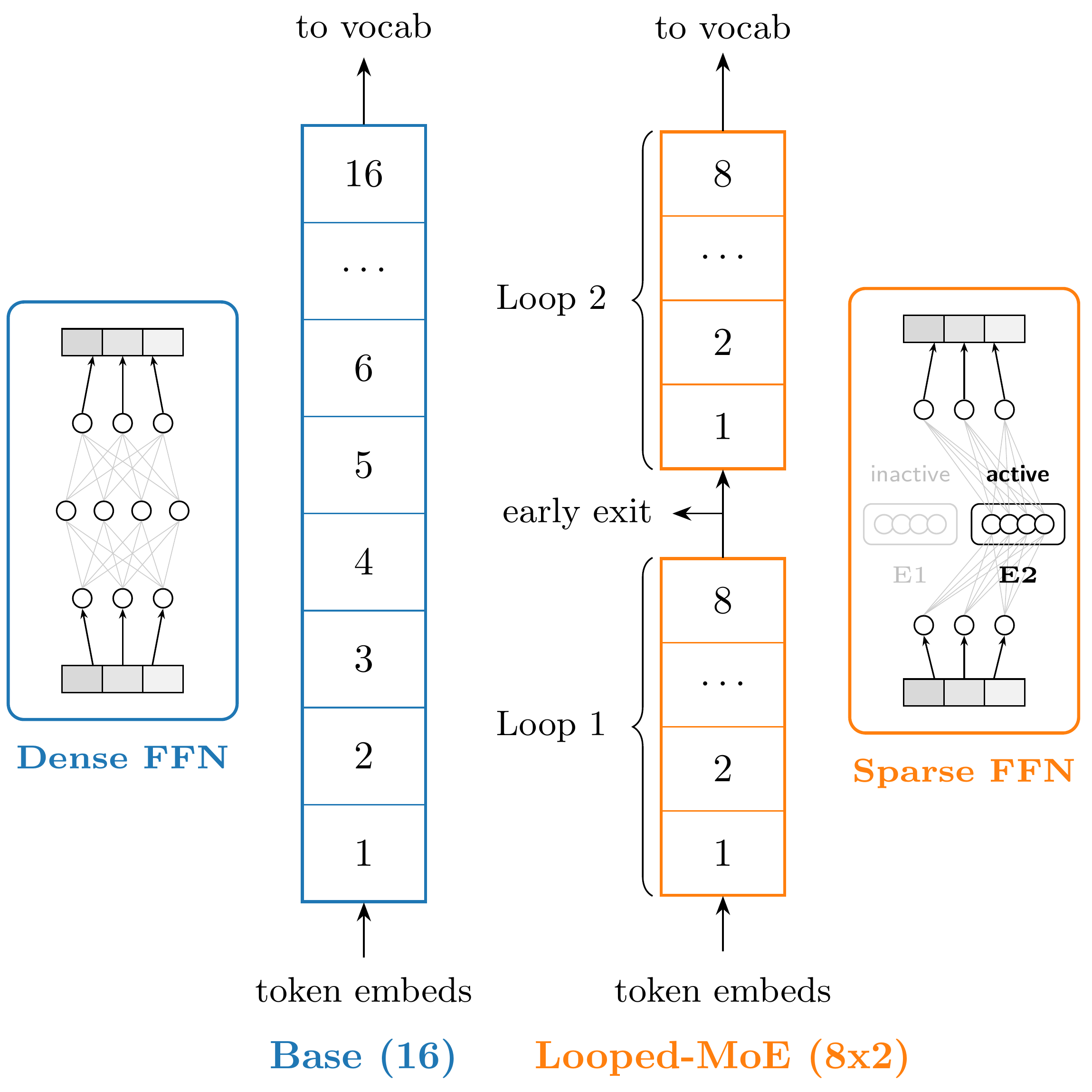}
        \label{fig:models}
    \end{minipage}%
    \hfill%
    \begin{minipage}[t]{0.65\linewidth}
        \centering
        \includegraphics[height=2.75in,width=\linewidth,keepaspectratio]{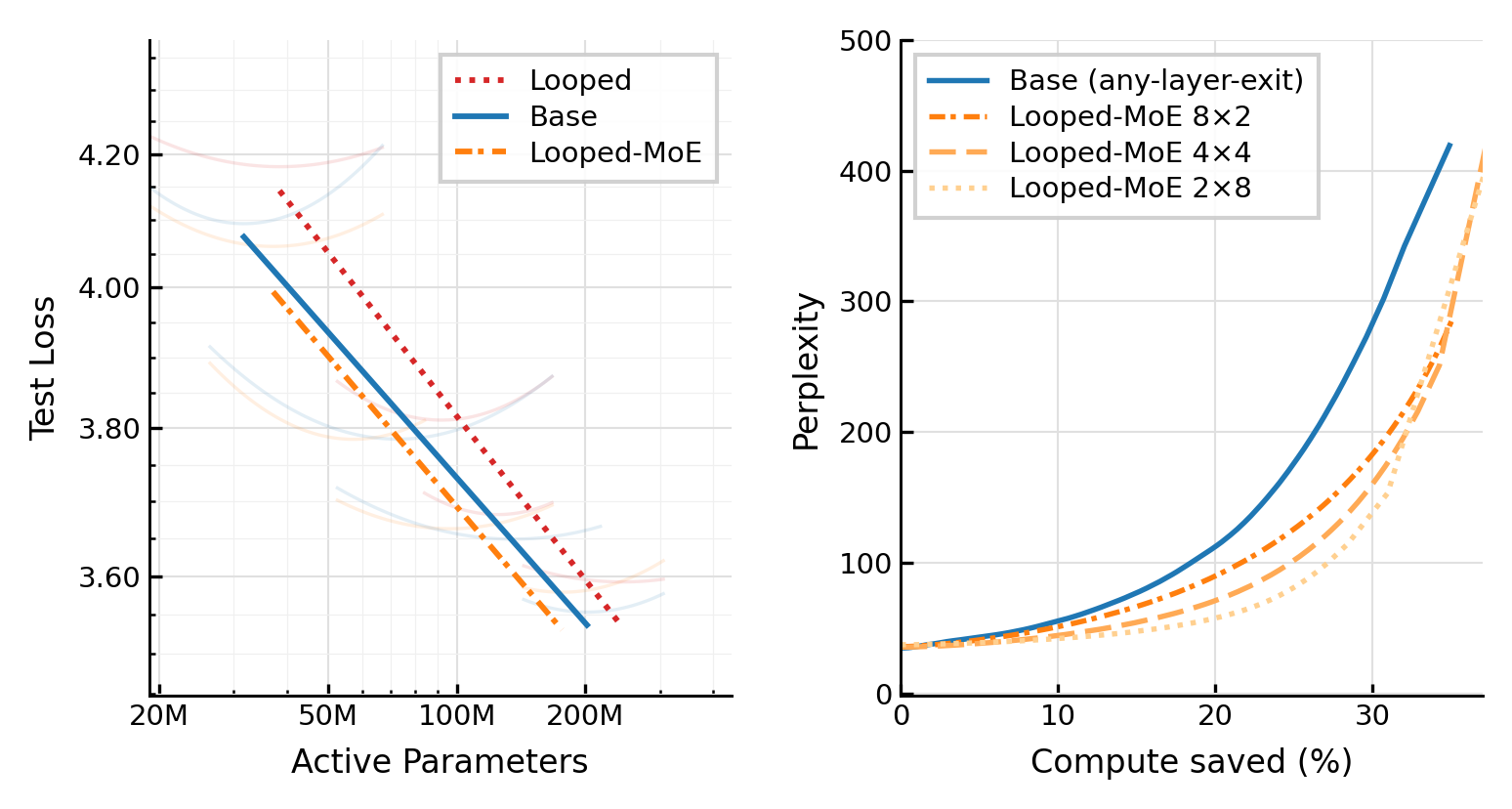}
        \label{fig:scaling-earlyexit}
    \end{minipage}
    \caption{Overview of our models and main results. \textbf{Left:} Base and Looped-MoE. Base models have unique layers and dense FFNs, while Looped-MoE models have repeated layer stacks with sparse MoE layers. \textbf{Middle:} IsoFLOP curves. Given the same number of parameters, Looped-MoE (8$\times$2) achieves lower test loss than Base across compute budgets. \textbf{Right:} Early-exit Pareto. Looped-MoE has a better compute-quality trade-off than Base models, and this improves with more looping.}
    \label{fig:overview}
\end{figure*}

\section{Model Descriptions}

The architectures compared in this work share a common backbone: a decoder-only transformer with multi-head self-attention (MHSA) using rotary positional embeddings (RoPE) \citep{su_roformer_2023}, SwiGLU feed-forward networks \citep{shazeer_glu_2020}, pre-RMSNorm \citep{zhang_root_2019}, and a residual stream (Appendix~\ref{app:compute}, Figure~\ref{fig:transformer_block}). To explore the impact of sparse layers on looped scaling laws, we vary model architecture by whether the FFN is dense or sparse and whether the layers are looped (Table~\ref{tab:architectures}). 

\begin{wraptable}{r}{0.4\textwidth}
\vskip -0.30in
\centering
\caption{Scaling Study Configurations.}
\label{tab:architectures}
\vskip -0.1in
\begin{tabular}{lcc}
\toprule
Architecture & FFN & Layers \\
\midrule
Looped   & Dense       & $8 \times 2$ \\
Base          & Dense       & 16 \\
Looped-MoE & Sparse  & $8 \times 2$ \\
MoE         & Sparse  & 16 \\ 
\bottomrule
\end{tabular}
\vskip -0.15in
\end{wraptable}

\subsection{Layer Looping}

As illustrated in Figure \ref{fig:overview} (left), in a looped transformer, a stack of $L$ unique transformer layers is repeated $R$ times in sequence during the forward pass, producing an effective depth of $L \times R$. The looped variants in the scaling study use $L{=}8, R{=}2$, matching the 16-layer effective depth of their non-looped counterparts. In Section \ref{subsec:loop-depth-early-exit}, we explore the impact of more looping with fewer layers. 

\subsection{Sparse Mixture-of-Experts}

In MoE, the SwiGLU FFN \citep{shazeer_glu_2020} with feed-forward dimension $d_\text{ff}$ is replaced with a set of experts (smaller SwiGLU FFNs). To select which expert FFNs to use per token, we use top-$k$ token-choice routing \citep{shazeer_outrageously_2017}: a learned router $h(x) = W_\text{router} x$ assigns each token embedding $x$ to its top-$k$ experts. Defining $W_\text{router} \in \mathbb{R}^{d_{\text{model}} \times E}$, where $E$ is the total number of experts, the forward pass from x to y is:
\[
\mathcal{T} = \text{top-}k(h(x)), \quad p_i(x) = \frac{e^{h(x)_i}}{\sum_{j \in \mathcal{T}} e^{h(x)_j}}, \quad y = \sum_{i \in \mathcal{T}} p_i(x)\, \mathrm{FFN}_i(x)
\]

To prevent expert collapse, we apply two standard MoE auxiliary losses: a load balancing loss \citep{fedus_switch_2022}, which encourages uniform token assignment across experts, and a router z-loss \citep{zoph_st-moe_2022}, which penalizes large logits to stabilize training:
\[
\mathcal{L}_{\text{LB}} = E \sum_{i=1}^{E} f_i \cdot \overline{p}_i,
\qquad
\mathcal{L}_{\text{RZ}} = \frac{1}{B} \sum_{j=1}^{B}
\left(
\log \sum_{i=1}^{E} \exp\!\left(h(x_j)_i\right)
\right)^{\!2}
\]
% \[
% \begin{aligned}
% \mathcal{L}_{\text{LB}} &= E \sum_{i=1}^{E} f_i \cdot \overline{p}_i,
% \qquad
% \mathcal{L}_{\text{RZ}} &= \frac{1}{B} \sum_{j=1}^{B}
% \left(
% \log \sum_{i=1}^{E} \exp(x_i^{(j)})
% \right)^2
% \end{aligned}
% \]
where $f_i$ is the fraction of tokens routed to expert $i$, $\overline{p}_i$ is its mean routing probability, $h(x_j)_i$ is the router logit for expert $i$ on token $j$, and $B$ is the number of tokens in the batch. We weight these losses by 0.01 ($\mathcal{L}_{\text{LB}}$) and 0.001 ($\mathcal{L}_{\text{RZ}}$), following OLMoE \citep{muennighoff_olmoe_2025}. For all MoE configurations in this study, we use $k{=}2$ active experts out of $E{=}8$ total, following Mixtral \citep{jiang_mixtral_2024}.

\section{Experimental Set-up}\label{sec:experimental-setup}

\subsection{Maximal Update Parameterization for Looped and Sparse Layers}
We use Maximal Update Parameterization ($\mu$P) \citep{yang_tensor_2022} to reduce the cost of our scaling studies. With $\mu$P, the optimal learning rate tuned at a small base width (we use $d_{\text{base}} = 128$) transfers to larger widths. This is efficient for isoFLOP studies, which require training models across a range of sizes: with $\mu$P, it is not necessary to re-tune the learning rate at every new width.

The core idea is to scale initialization variance and learning rates inversely with a width ratio $w_{\text{ratio}} = d_{\text{model}} / d_{\text{base}}$, so that activation and update magnitudes remain consistent across model scales. We apply standard $\mu$P to all non-embedding weight matrices (Table~\ref{tab:mup}) then extend these scalings to the unembedding matrix $W_{\text{unembed}}$ in place of the output multiplier used in the original formulation.

We treat MoE expert weights and the router as additional hidden weights and apply width-scaled initialization and learning rates. Unlike prior approaches~\citep{malasnicki_-parametrization_2025}, we find it sufficient to only use two scaling mechanisms, initialization variance and per-layer learning rate (without additional forward-pass multipliers). For looped layers, no additional modification is needed: weight-tied layers receive gradient contributions from multiple positions but their scale is unchanged.

\begin{figure}[t]
    \centering
    \includegraphics[width=\textwidth]{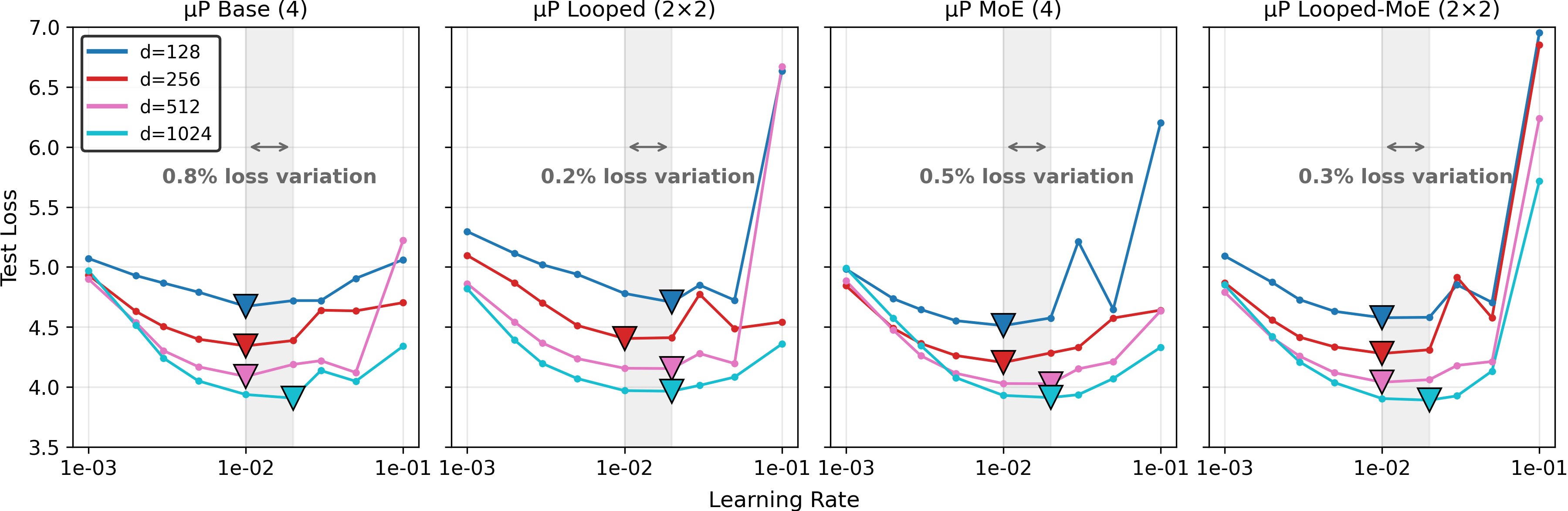}
    \caption{$\mu$P Transfer test. The best learning rate for the smallest model remains optimal across larger sizes, validating our $\mu$P implementation. If the optimal learning rate for a given width does differ from the base learning rate, the loss difference is < 1\%.}
    \label{fig:mu_transfer}
\end{figure}

\begin{table}[t]
\centering
\caption{$\mu$P scaling rules for all weights except the embedding ($w_{\text{ratio}} = d_{\text{model}} / d_{\text{base}}$). We extend these to router, expert, and unembedding weights, removing the original input/output multipliers.}
\label{tab:mup}
\begin{tabular}{ccc}
\toprule
\textbf{Init.\ Variance} & \textbf{Learning Rate} & \textbf{New for MoE, Looped} \\
\midrule
$\sigma_{\text{base}}^2 / w_{\text{ratio}}$ & $\eta_{\text{base}} / w_{\text{ratio}}$ & $W_{\text{unembed}}$, $W_{\text{router}}$, $W_{\text{experts}}$ \\
\bottomrule
\end{tabular}
\end{table}

We validate our parameterization with a $\mu$P transfer test on 4-layer effective depth models. We find that the best learning rate at the smallest width is also near-optimal for the larger widths ($d_\text{model} \in \{128, 256, 512, 1024\}$), with at most 0.8\% loss difference across all architectures \mbox{(Figure \ref{fig:mu_transfer})}.
\subsection{FLOPs Accounting for Looped and Sparse Layers}
For calculating compute, we follow the $C = 6ND$ approximation \citep{kaplan_scaling_2020,hoffmann_training_2022}, where $C$ is a compute budget in FLOPs, $N$ is the number of total parameters, and $D$ is data in tokens. Here, we use $N_\text{active}$, not $N_\text{unique}$ as $N$ in the equation. Concretely, we define $N_{\text{active}}$ as the total parameters \textit{used} in a single forward pass, counting repeated (looped) layers at each invocation and counting only the $k$ active experts in MoE layers. This is distinct from $N_{\text{unique}}$, the number of stored parameters as described in Table~\ref{tab:flops}. By design, all four architectures have the same $N_{\text{active}}$ parameters (neglecting the small router weights) and therefore the same compute budget at matched token counts. We include embedding and unembedding parameters in our $N_\text{active}$ count for compute calculations, an additional $(2  \cdot V \cdot d_\text{model}$) parameters, where $V$ is vocabulary size.

\begin{table}[t]
\centering
\caption{Active vs.\ stored non-embedding parameters. In this table we compare the number of parameters used in the forward pass ($N_\text{active}$) against the number of parameters stored ($N_\text{unique}$), accounting for inactive expert weights in MoE. $L$ is effective layers (counting repeated layers), $A$ is attention parameters per layer, $F$ is FFN parameters per layer, and $F_{\text{expert}} = F / k$ is parameters per expert for $k$ active experts. In our total $N_\text{active}$, we also include embedding parameters (not shown).}
\label{tab:flops}
\begin{tabular}{l r @{\;} c @{\;} l}
\toprule
Architecture & \multicolumn{1}{c}{$N_{\text{active}}$} & \multicolumn{2}{c}{\hspace{-2.5em}$N_{\text{unique}}$} \\
\midrule
Looped ($R$ loops) & $L(A + F)$ & $>$ & $\frac{L}{R}(A + F)$ \\
Base & $L(A + F)$ & $=$ & $L(A + F)$ \\
Looped-MoE ($R$ loops) & $L(A + F)$ & $<$ & $\frac{L}{R}(A + E \cdot F_{\text{expert}})$ \\
MoE ($E$ experts, $k$ active) & $L(A + F)$ & $\ll$ & $L(A + E \cdot F_{\text{expert}})$ \\
\bottomrule
\end{tabular}
\end{table}

%% ============================================================
\section{Experiments}\label{sec:experiments}
%% ============================================================

\subsection{IsoFLOP Scaling Study}
We find scaling laws \citep{hoffmann_training_2022} for each of our four architectures to compare them. Given a fixed compute budget $C$, we train models at several widths (Appendix~\ref{app:compute}, Table~\ref{tab:model_configs}). For each width, the number of training tokens is determined by $D = C / (6 N_{\text{active}})$, so that all runs at a given budget use the same total compute. Each run produces a test loss at its $(N, D)$ configuration; we fit a quadratic to these losses and take the minimum as the compute-optimal model for that budget. Repeating this across four budgets $C \in \{5 \times 10^{16},\; 2 \times 10^{17},\; 5 \times 10^{17},\; 10^{18}\}$ FLOPs yields a set of compute-optimal points, through which we fit a power law for loss vs.\ parameters, $L \propto N^{-\alpha}$.

We pretrain on the 10B token sample of FineWeb-Edu \citep{penedo_fineweb_2024} using the GPT-2 tokenizer ($V = 50{,}257$), with AdamW ($\beta_1 = 0.9$, $\beta_2 = 0.999$, $\epsilon = 10^{-8}$, independent weight decay $1.0 \times 10^{-4}$). We use a Warmup-Stable-Decay (WSD) learning rate schedule \citep{hu_minicpm_2024} with sqrt-decay cooldown \citep{dremov_training_2025} to $5\%$ of peak over the final $10\%$ of training steps. The peak learning rate is determined via $\mu$Transfer from the $d_{\text{base}} = 128$ proxy model.  We provide additional hardware and train time details in Appendix~\ref{app:compute}.

% We split our pretraining dataset into train and test, and report our test losses on the unseen split
To test whether our hypothesis, that sparse routing restores the expressiveness lost to weight tying, holds beyond test loss, we evaluate the compute-optimal Base, Looped, and Looped-MoE models at our largest compute budget on the AI2 OLMES benchmark suite \citep{gu_olmes_2025}.

\subsection{Early Exit Evaluation}\label{subsec:early-exit}
To understand how looped models can save inference time in addition to memory, we evaluate the theoretical compute savings of early exit using training-free criteria. We measure the compute-quality trade-off using entropy of the output distribution \citep{teerapittayanon_branchynet_2016, xin_deebert_2020} as the exit criterion. As shown in Figure \ref{fig:early-exit-overview} (left), at each candidate exit point, we project the hidden state to the vocabulary and exit if the entropy falls below a threshold $\tau$. If tokens exit early, we count the unused depth as FLOPs saved.

For looped models, exit is permitted only at loop boundaries after each complete pass through the unique layer stack. For non-looped models, exit is permitted at any intermediate layer. We sweep $\tau$ to trace a Pareto frontier of compute saved (as a percentage of the full forward pass) versus perplexity. We do not measure end-to-end inference throughput. All early exit experiments use compute-optimal models at $10^{18}$ training FLOPs. Besides entropy, we also test a confidence threshold (Appendix~\ref{app:compute}).

\section{Results}

\subsection{IsoFLOP Scaling Laws}
\begin{figure}[t]
    \centering
    \begin{subfigure}{0.48\linewidth}
        \centering
        \includegraphics[width=\linewidth]{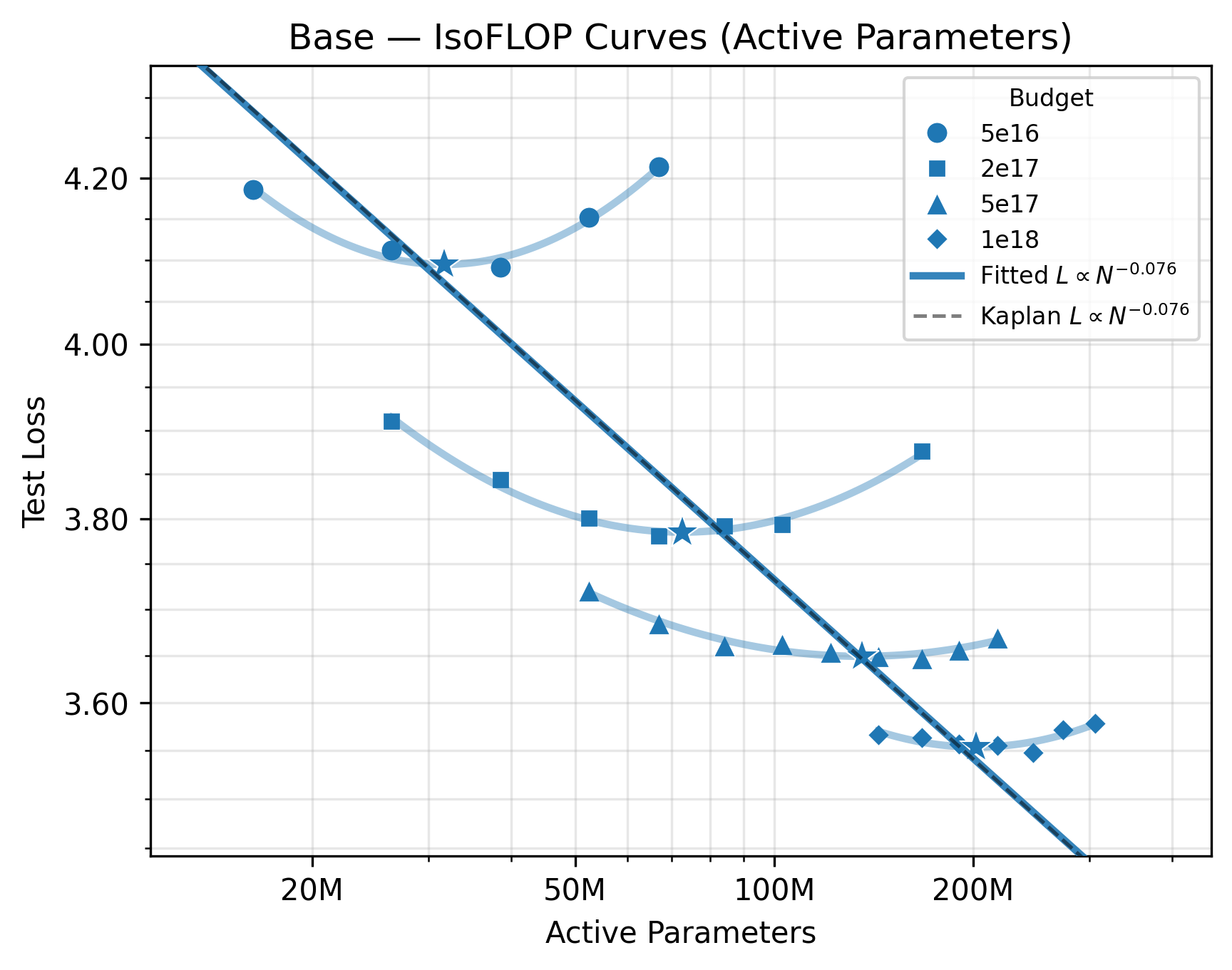}
    \end{subfigure}
    \hfill
    \begin{subfigure}{0.48\linewidth}
        \centering
        \includegraphics[width=\linewidth]{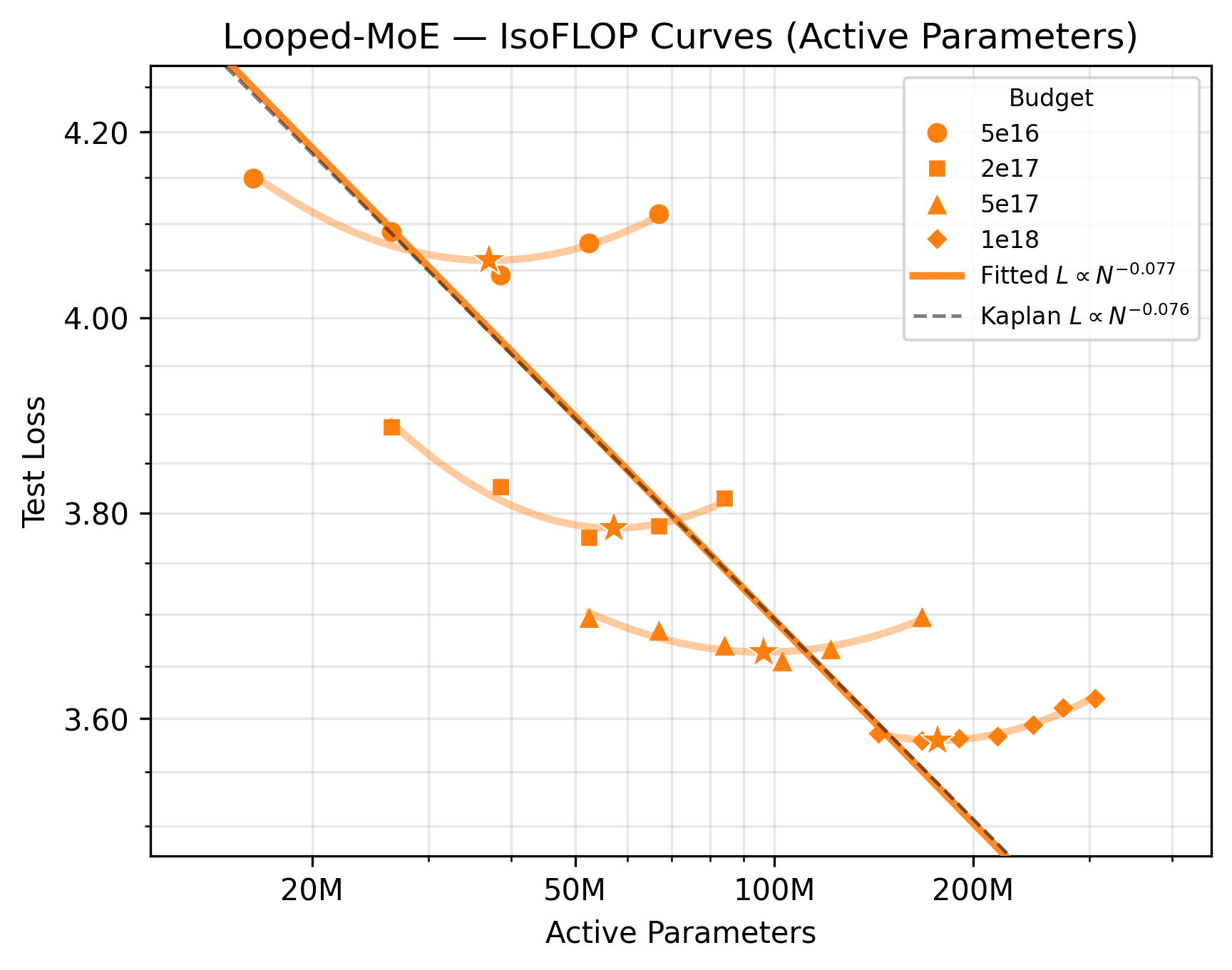}
    \end{subfigure}
    \caption{\textbf{Left:} IsoFLOP curves for Base. \textbf{Right:} IsoFLOP curves for Looped-MoE. Stars mark compute-optimal model sizes at each budget; solid lines show fitted $L \propto N^{-\alpha}$ ($\alpha = 0.076$ for Base, $0.077$ for Looped-MoE). The dashed line shows \citet{kaplan_scaling_2020} scaling exponent ($\alpha = 0.076$).}
    \label{fig:isoflops_individual}
\end{figure}

We find that Looped-MoE scales better than Base, while Looped models scale strictly worse (Figure~\ref{fig:overview}, middle), supporting our hypothesis that sparse routing restores expressiveness lost to weight tying. Looping alone is detrimental, with MoE scaling better than Looped-MoE and Base better than Looped (Appendix~\ref{app:compute}, Figure~\ref{fig:isoflops_combined_all}), consistent with prior work~\citep{kaplan_scaling_2020, prairie_parcae_2026}. Yet looped models remain appealing for their memory savings and early-exit points, and we show that sparse layers close the scaling gap. Figure~\ref{fig:isoflops_individual} shows the individual isoFLOP curves for Base and Looped-MoE; fits for Looped and MoE are in Appendix~\ref{app:compute}, Figure~\ref{fig:isoflops_individual_others}. Fitting a power law for Loss, $L \propto N^{-\alpha}$ through the compute-optimal points yields $\alpha = 0.076$ (Base) and $0.077$ (Looped-MoE), consistent with the $0.076$ scaling exponent reported by \citet{kaplan_scaling_2020}.

\subsection{Downstream Evaluation}
The downstream results also confirm finding (i): sparse experts resolve the looped scaling deficit. Table~\ref{tab:bench_1e18} reports AI2 OLMES benchmark results for the compute-optimal Base, Looped, and Looped-MoE models at the $10^{18}$ FLOP budget over 6 seeds. Looped-MoE achieves the highest average score (37.1), above Base (36.9) despite storing fewer unique parameters (216M vs.\ 246M). We exclude BoolQ due to its high variance across seeds; full results including BoolQ and MoE are in Appendix~\ref{app:compute}, Table~\ref{tab:bench_1e18_full}.

\begin{table}[t]
  \centering
  \caption{AI2 OLMES benchmark results for compute-optimal models at $10^{18}$ FLOPs (mean $\pm$ std, 6 seeds). Looped-MoE achieves the highest average score with fewer stored parameters than Base.}
  \label{tab:bench_1e18}
  \resizebox{\linewidth}{!}{%
  \begin{tabular}{llcccccccc|c}
    \toprule
    \textbf{Model} & \textbf{Params} & \textbf{ARC-E} & \textbf{ARC-C} & \textbf{CSQA} & \textbf{HellaSwag} & \textbf{OBQA} & \textbf{PIQA} & \textbf{SIQA} & \textbf{WinoG} & \textbf{Avg.} \\
    \midrule
    Looped & 168M & 41.1\,$\pm$\,0.5 & 24.4\,$\pm$\,1.2 & \textbf{30.0}\,$\pm$\,1.6 & 25.5\,$\pm$\,0.5 & 26.5\,$\pm$\,1.6 & 56.6\,$\pm$\,0.5 & \textbf{39.8}\,$\pm$\,0.7 & 51.2\,$\pm$\,1.0 & 36.9\,$\pm$\,0.3 \\
    Base & 246M & \textbf{41.6}\,$\pm$\,1.3 & \textbf{24.7}\,$\pm$\,1.0 & 28.5\,$\pm$\,1.0 & \textbf{26.3}\,$\pm$\,0.8 & \textbf{27.1}\,$\pm$\,1.7 & \textbf{56.8}\,$\pm$\,0.7 & 39.5\,$\pm$\,0.6 & 51.2\,$\pm$\,0.8 & 36.9\,$\pm$\,0.4 \\
    Looped-MoE & 216M & 41.4\,$\pm$\,0.9 & 24.4\,$\pm$\,0.4 & 29.8\,$\pm$\,1.2 & 26.1\,$\pm$\,0.4 & 26.7\,$\pm$\,1.4 & 56.5\,$\pm$\,0.6 & 39.6\,$\pm$\,0.8 & \textbf{52.3}\,$\pm$\,0.9 & \textbf{37.1}\,$\pm$\,0.4 \\
    \bottomrule
  \end{tabular}%
  }
\end{table}

\subsection{Early Exit}
\begin{figure*}[t]
    \centering
    \begin{minipage}[t]{0.32\linewidth}
        \centering
        \includegraphics[height=2.75in,width=\linewidth,keepaspectratio]{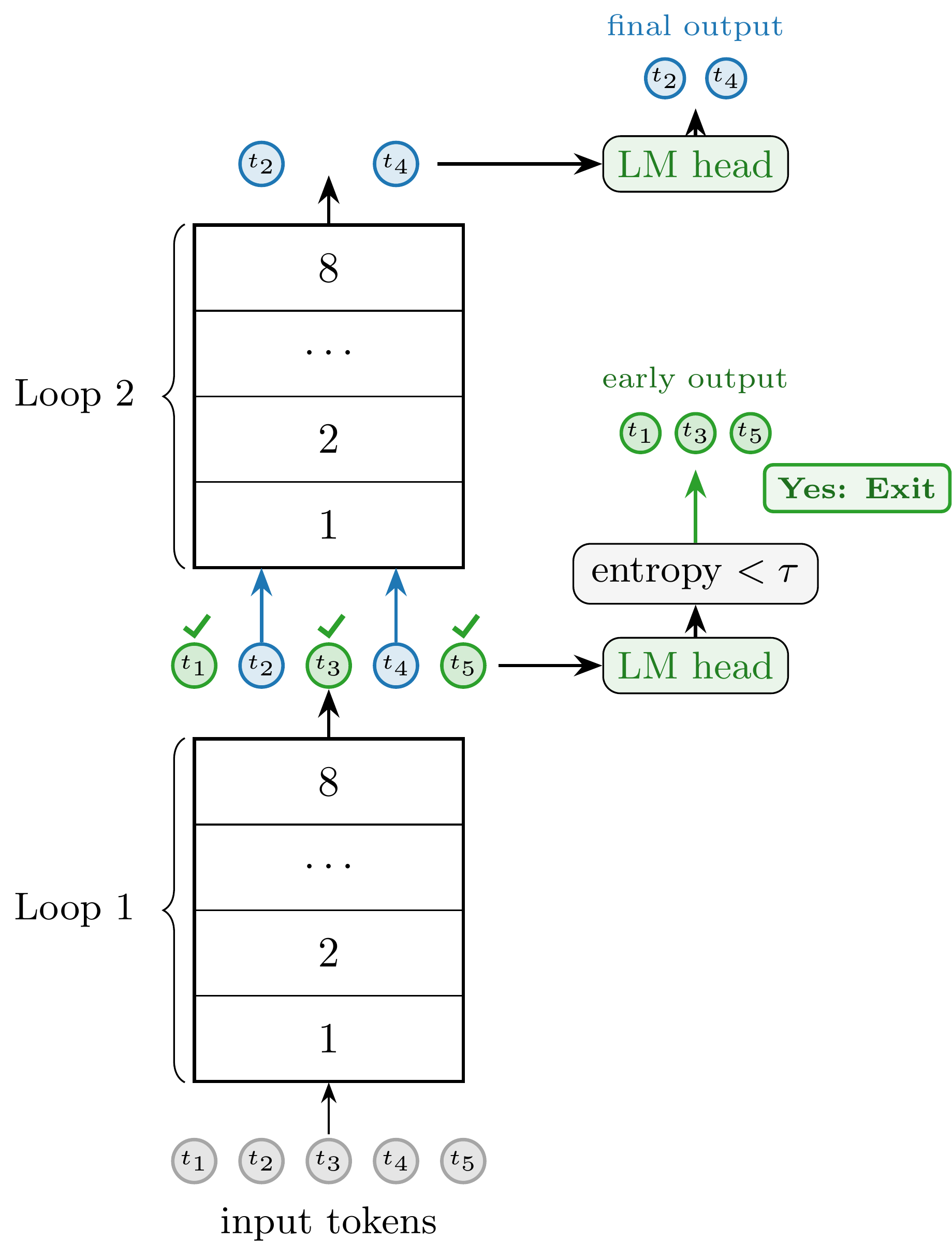}
    \end{minipage}%
    \hfill%
    \begin{minipage}[t]{0.66\linewidth}
        \centering
        \includegraphics[height=2.75in,width=\linewidth,keepaspectratio]{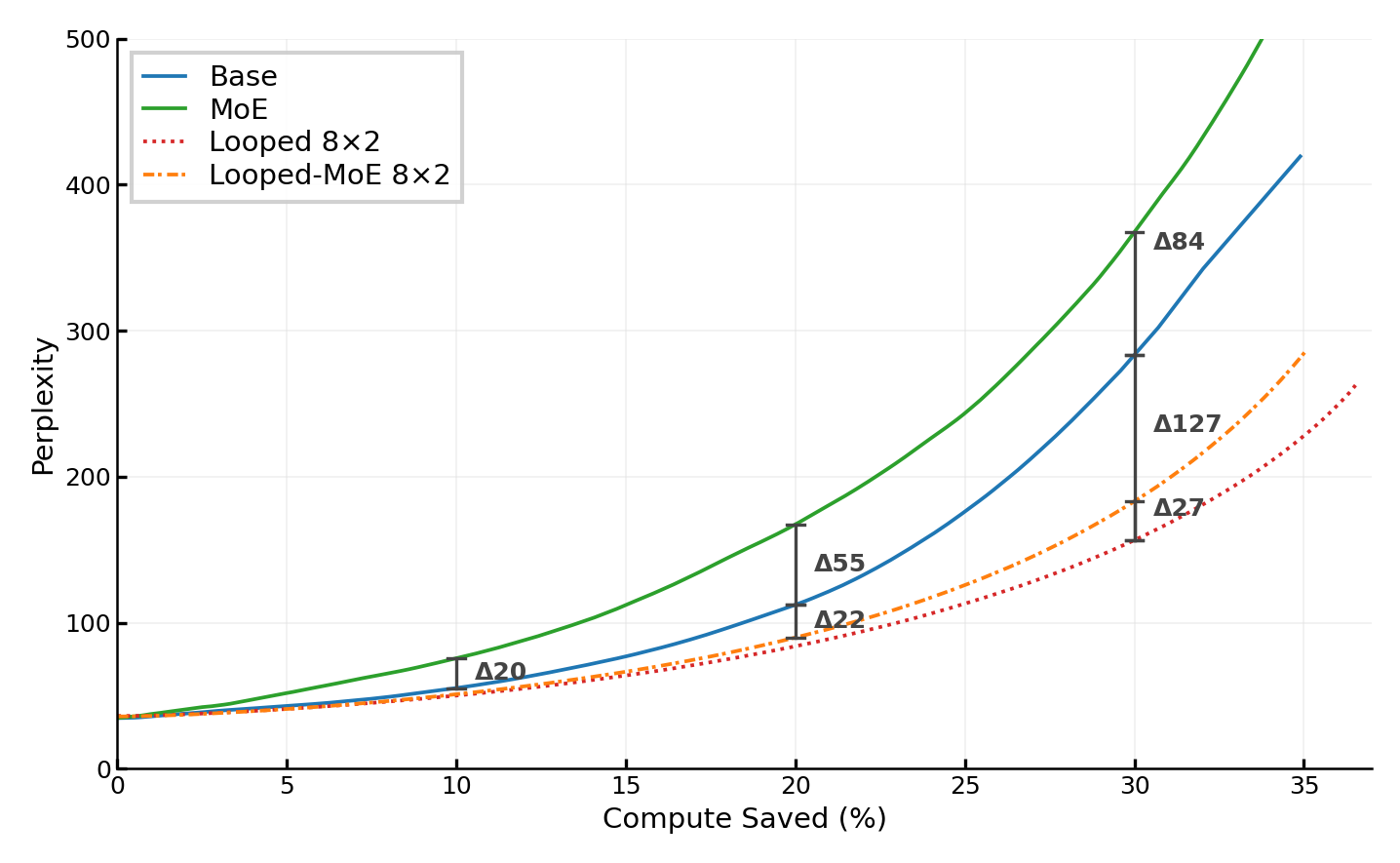}
    \end{minipage}
    \caption{\textbf{Left:} Schematic of entropy-based early exit. For Looped/Looped-MoE, tokens exit at loop boundaries. \textbf{Right:} Looped/Looped-MoE models have the best compute-quality trade-offs.}
    \label{fig:early-exit-overview}
\end{figure*}

Figure~\ref{fig:early-exit-overview} shows compute-quality trade-off under a training-free entropy criterion; Appendix~\ref{app:compute}, Table~\ref{tab:ee_perplexity} reports it in tabular format. Looped models degrade less than their non-looped counterparts: at 10\% FLOPs saved, Looped and Looped-MoE reach perplexity 50.2 and 51.0, compared to 55.4 for Base and 75.7 for MoE. Notably, MoE degrades fastest, suggesting that sparse routing is not the primary contributor to favorable early-exit trade-offs. We hypothesize the benefit comes from looping itself: the last layers that produce the final output are also the last layers in each loop, and investigate this in Section~\ref{subsec:convergence}.

\section{Analysis}
In this section, we conduct experiments to understand why replacing the dense FFN of a looped transformer with a MoE layer results in better scaling laws and early-exit trade-offs. At a high level we find the reasons are: (1) loop-specific expert assignments (Sec \ref{subsec:routing}) and (2) earlier convergence to final activations at loop points, due to the final layers at the end of loops being the same layers trained to produce the final activations (Sec \ref{subsec:convergence}). For Looped-MoE, we find that more looping further improves early-exit trade-offs (Sec \ref{subsec:loop-depth-early-exit}); we leave the impact on scaling for future work. All experiments use models at $10^{18}$ training FLOPs, evaluated on 800K test tokens.

\subsection{Why do MoE Layers Improve Looped Scaling Laws?}
\label{subsec:routing}

Our scaling laws and downstream evaluations reveal that Looped-MoE architectures consistently outperform dense baselines at equivalent compute budgets. We hypothesize that MoE routing overcomes the expressive limitation of weight tying by activating different expert sub-networks on each loop pass. If we are correct, when the same token passes through the same physical layer on the second loop iteration, the router should assign it to different experts. If expert assignments diverge across loops, then MoE layers effectively specialize by loop iteration and the same layers have depth-unique computations.

\paragraph{Setup.} For the compute-optimal Looped-MoE model at $10^{18}$ training FLOPs, we track the top-$k$ expert assignments on both loop passes. With $k=2$ active experts out of $E=8$ total, we record whether each token's expert set on pass~2 fully overlaps, partially overlaps, or is entirely disjoint from its pass~1 assignment.

\begin{figure}[t]
    \centering
    \includegraphics[width=0.8\linewidth]{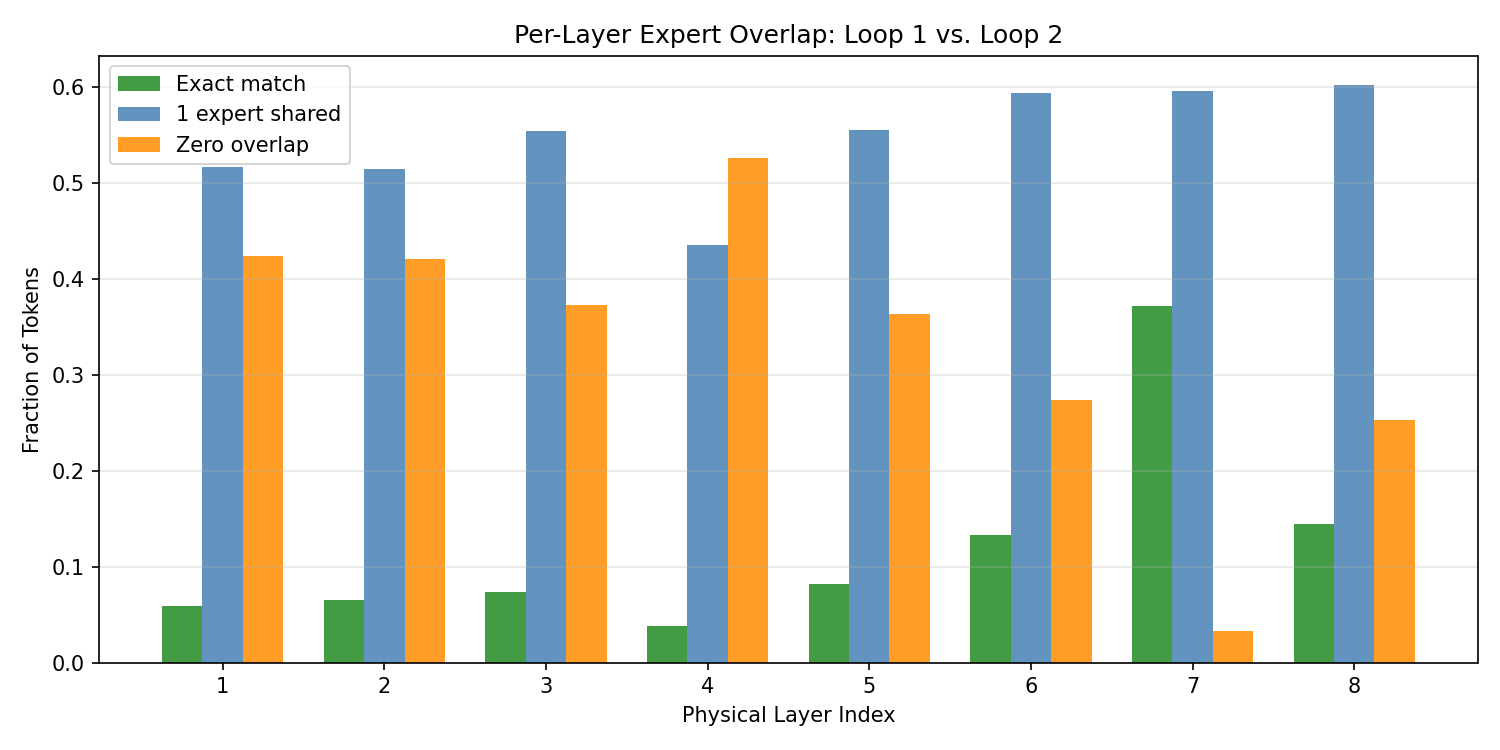}
    \caption{Expert assignment overlap between loop passes 1 and 2 across physical layers in a Looped-MoE model (8 layers $\times$ 2 loops, $k=2$, $E=8$). The majority of tokens receive a unique set of experts across loops, with layer~7 as a notable exception exhibiting high routing consistency.}
    \label{fig:routing-divergence}
\end{figure}

\paragraph{Routing predominantly diverges across loops.} Figure~\ref{fig:routing-divergence} presents the per-layer breakdown of expert assignment overlap. Across layers 1--6 and 8, 25--53\% of tokens receive entirely non-overlapping expert assignments between passes, while only 4--14\% receive identical assignments. For a majority of tokens, exactly one of the two active experts is shared between loops---the other changes. This demonstrates that MoE routing enables shared layers to deploy substantially different expert sub-networks on each loop iteration for the majority of tokens.

\paragraph{Layer-specific behavior.} Layer~7 is a notable outlier, exhibiting markedly higher overlap (37\% exact match and only 3\% zero overlap). We speculate this loop-invariance reflects a structural role for layer~7, possibly stabilizing embeddings before vocabulary projection at the loop boundary. Overall, we observe unique computations per loop, unlike the fixed computation of a dense repeated layer.

\subsection{Why are Early Exits for Looped Transformers Favorable?} \label{subsec:convergence}
In looped models the layers at each exit point are the same layers that produce the final output projected to vocabulary space. We hypothesize that this architectural property results in loop boundary activations being better formed for vocabulary projection and thus early exiting. We test this by measuring how similar each layer's output distribution is to the final output, expecting looped models to be closer to final distributions at loop boundaries.

\paragraph{Setup.} We evaluate the compute-optimal model for each architecture at the $10^{18}$ FLOPs budget on our test tokens. We capture hidden states at every effective layer during inference and project each through the final layer norm and language model head~\citep{nostalgebraist_interpreting_2020}. We then compute the Jensen-Shannon divergence (JSD) between each intermediate output distribution and the final layer's output distribution. Concretely, for each token at effective layer $\ell$:
\begin{equation}
    \text{JSD}(p_\ell \| p_L) = \frac{1}{2} D_{\text{KL}}(p_\ell \| m) + \frac{1}{2} D_{\text{KL}}(p_L \| m), \quad m = \frac{1}{2}(p_\ell + p_L)
    \label{eq:jsd}
\end{equation}
where $p_\ell$ is the distribution at layer $\ell$, $p_L$ is the final layer's distribution, and $m$ is their average. We normalize JSD to $[0, 1]$ and consider a token converged if its normalized JSD falls below $0.5$.

\begin{figure}[t]
    \centering
    \includegraphics[width=\linewidth]{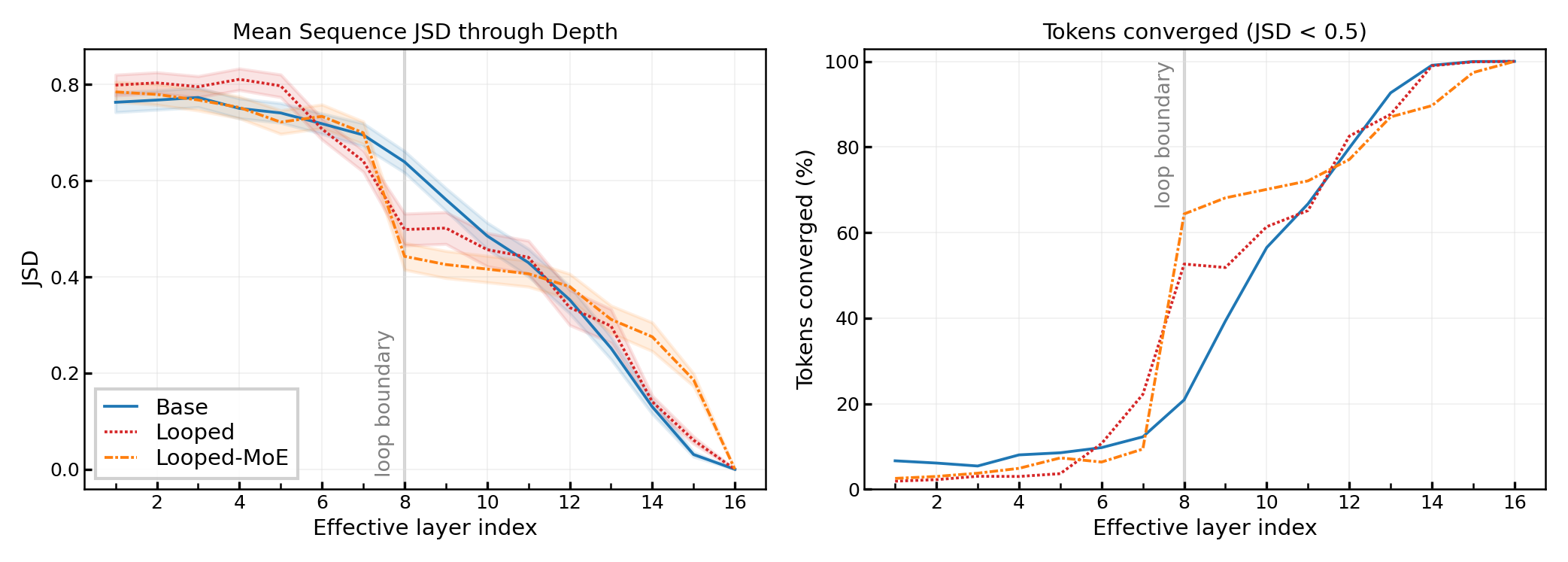}
    \caption{Distributional analysis of intermediate layer outputs relative to the final layer output (JSD), using compute-optimal models at $10^{18}$ FLOPs. \textbf{Left:} Mean JSD at each effective layer. Shaded bands show standard deviation of JSD across sequences. \textbf{Right:} Fraction of tokens at a given effective layer with JSD $< 0.5$. Looped models converge substantially at loop boundaries, supporting our hypothesis that shared exit layers produce well-formed outputs.}
    \label{fig:convergence}
\end{figure}

\paragraph{Looped models converge faster at loop boundaries.} Figure~\ref{fig:convergence} shows the mean JSD and fraction of converged tokens (JSD $< 0.5$) at each effective layer. Looped models converge a substantially larger fraction of tokens at earlier depths than Base, with a sharp jump at the loop boundary. By the end of the first loop iteration, the majority of tokens have already reached near-final output distributions. Looped-MoE converges more tokens than Looped, which we attribute to more expressive MoE layers; yet expressivity alone is insufficient, as MoE converges only $\sim$30\% of tokens at layer 8 vs.\ $\sim$65\% for Looped-MoE.

\paragraph{The mechanism is architectural, not learned.} In a looped model, every loop iteration ends with the same layers that produce the final output. An early exit at the end of loop 1 projects through layers that were trained as the model's output-facing computation. In a non-looped model, an exit at the equivalent depth (layer 8 of 16) projects through layers that were trained as intermediate representations, never optimized to produce well-formed output distributions. This property requires no additional training signal---it is an inherent consequence of layer looping.

\begin{figure}
    \centering
    \includegraphics[width=\linewidth]{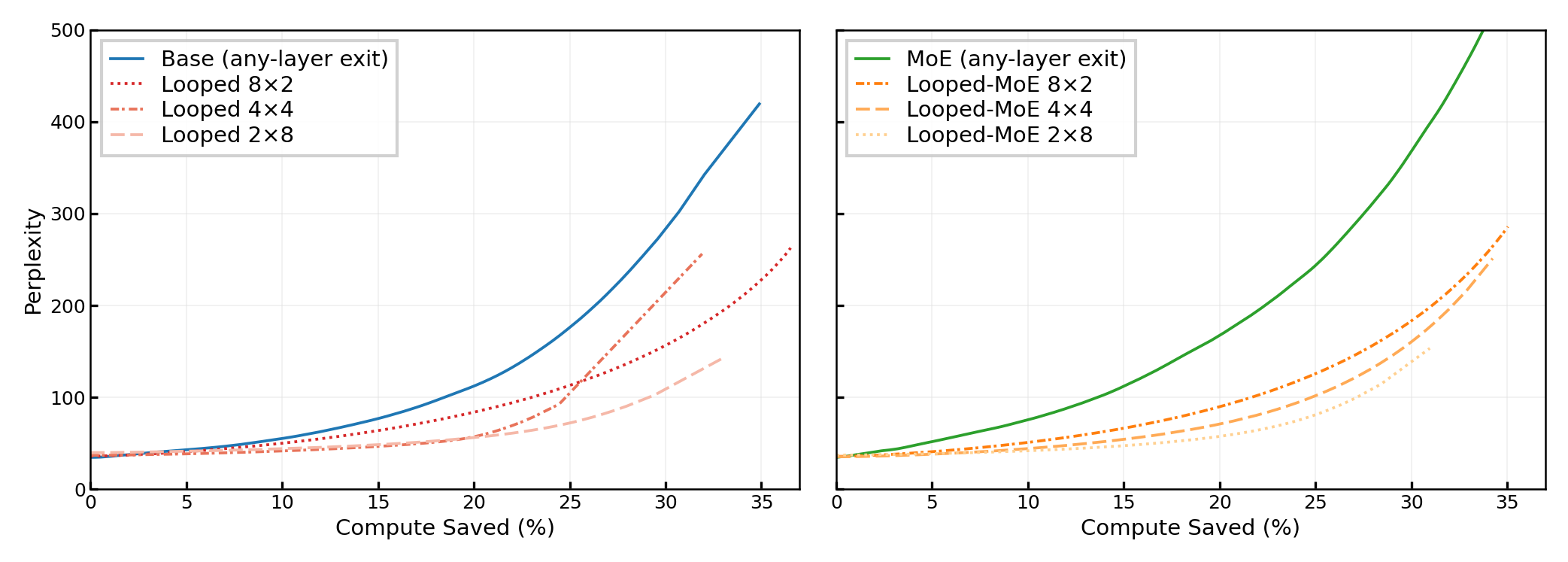}
    \caption{\textbf{Left:} More loops improve the Looped compute-quality trade-off, though not strictly at all savings levels. \textbf{Right:} For Looped-MoE, more loops generally yield a better compute-quality trade-off, with all configurations better than non-looped MoE.}
    \label{fig:early-exit-looping}
\end{figure}

\subsection{Does More Looping Improve Early-Exit Efficiency?}
\label{subsec:loop-depth-early-exit}

The looped architecture introduces natural early-exit points at loop boundaries---positions where a full pass through the shared layer stack has completed. A model with $L$ physical layers and $R$ loops has $R-1$ such exit points: 1 for $8 \times 2$, 3 for $4 \times 4$, and 7 for $2 \times 8$. We ask: does increasing the number of loops (and thus exit points) improve the compute-quality trade-off?

\paragraph{Setup.} We train Looped and Looped-MoE variants at the $10^{18}$ FLOPs budget with three loop-depth configurations: $8 \times 2$ (8 layers, 2 loops), $4 \times 4$ (4 layers, 4 loops), and $2 \times 8$ (2 layers, 8 loops). All three share the width of the compute-optimal $8 \times 2$ configuration ($d_\text{model}=896$ for Looped, $704$ for Looped-MoE) and have 16 effective layers. We apply the same entropy-based early-exit procedure described in Section~\ref{subsec:early-exit}, with tokens exiting at the first loop boundary where entropy falls below $\tau$.

\paragraph{More loops improve the early-exit trade-off.} Increasing the number of loops generally yields a better compute-quality trade-off, as reported in Figure~\ref{fig:early-exit-looping}. For Looped-MoE, $2 \times 8$ has the best compute-quality trade-off at most savings levels, followed by $4 \times 4$, then $8 \times 2$. This improvement arises from two factors: (1)~more loop boundaries provide more opportunities to exit where the output is sufficiently converged; and (2)~every loop boundary is a high-quality exit point, since the token has passed through the same layers that produce the final output.

\paragraph{Looped-MoE outperforms non-looped MoE.} All three Looped-MoE configurations achieve a better early-exit Pareto frontier than the non-looped MoE baseline, despite the non-looped MoE having access to 15 candidate exit layers with no restriction on which layer a token may exit at. The looped variants, by contrast, are restricted to exiting only at loop boundaries. Even with this structural disadvantage, looped models offer a better compute-quality trade-off.

\section{Related Works}
\paragraph{Looped Transformer Architectures.} MoEUT \citep{csordas_moeut_nodate} extends the looping Universal Transformer with $\sigma$-MoE \citep{csordas_approximating_2023} but also includes SwitchHead \citep{csordas_switchhead_2024} and a modified LayerNorm \citep{ba_layer_2016} as part of a broader re-design. We focus on the standard token-choice top-k MoE, and specifically choose to only change this from baseline dense looped model, to investigate whether dense FFNs are the expressive bottleneck for looped architectures. Ouro \citep{zhu_scaling_2025} is a family of 1.4B and 2.6B looped transformers; however, they loop dense FFNs, which we have shown are sub-optimal when truly compared to dense baselines on fixed compute. In their study, they compare performance by the number of unique parameters, while giving their models much more compute ($4\times$ the effective depth) than the models in their comparisons. In contrast, we conduct a comparative study in the isoFLOP setting: understanding instead when compute is fixed, which model architectures scale better. Concurrently, Parcae \citep{prairie_parcae_2026} stabilizes looped training via spectral norm constraints and establishes looping as an orthogonal scaling axis to data. Our work is complementary, focusing on the expressivity bottleneck of dense looped layers and early-exit efficiency.

\paragraph{Early Exits and Layer Skipping.} Early exits have been explored extensively for deep neural networks and more recently for transformers. LayerSkip \citep{elhoushi_layerskip_2024} trains standard transformers with layer dropout and a shared exit across all layers, enabling early-exit inference. In contrast, we observe looped architectures possess this property by construction: the same stack of layers already handles final activations during normal training. Mixture of Depths \citep{raposo_mixture--depths_2024} dynamically routes tokens to skip layers via a learned top-k mechanism, and Mixture of Recursions \citep{bae_mixture--recursions_2025} extends this to looped architectures by learning per-token recursive depths. Like MoEUT, these works demonstrate favorable scaling for compute-adaptive models, but involve substantially more complex deviations from standard architectures. Our early-exit analysis instead offers a focused study of what enables looped models to scale: we find that switching to sparse MoE layers is the key step change, and that the resulting convergence properties naturally support early exiting as a downstream benefit rather than as a separate architectural modification.

\section{Conclusion and Limitations}
\paragraph{Conclusion.} We present a controlled study comparing dense, MoE, looped, and Looped-MoE transformers, isolating the effect of sparse expert layers on looped model scaling and downstream performance. Our results show that replacing dense FFNs with top-k MoE layers resolves the scaling deficit of looped transformers. We provide a mechanistic explanation for Looped-MoE's advantage: MoE routing diverges across loop iterations, allowing shared layers to implement distinct computations per loop. We also find that looping drives earlier output convergence at loop boundaries, enabling training-free early exits with better compute-quality trade-offs. Together, these findings suggest that Looped-MoE models are a practical path toward language models that are cheaper to store, faster to run, and competitive in quality.

\paragraph{Limitations.} $\mu$P transfer does not hold when model depth changes, so for our main experiments we hold depth constant at 16 effective layers and scale width. To check generalization to deeper models, we repeat our comparison at 32 effective layers, where our main findings hold (Appendix~\ref{app:32L}). Future work could validate with depth-scaling extensions such as CompleteP \citep{dey_dont_2026}. Due to compute constraints, we did not scale beyond 305M/711M (active/stored) parameters, but rely on compute-optimal scaling laws fitted at smaller scales to predict larger model performance. We aim to validate this with extended pretraining at 1B and 7B scales. Finally, our early-exit savings are theoretical; we have not measured end-to-end throughput with an optimized inference engine.

\section*{Acknowledgments}
Approved for public release; distribution is unlimited. This material is based upon work supported by the Defense Advanced Research Projects Agency (DARPA) under Agreement No. HR00112590089. The Authors acknowledge the National Artificial Intelligence Research Resource (NAIRR) Pilot, the Texas Advanced Computing Center (TACC) at The University of Texas at Austin, and the Jetstream2 cloud resource supported by the National Science Foundation (award NSF-OAC 2005506) at Indiana University for providing computational resources that have contributed to this research result.

\newpage
\bibliographystyle{unsrtnat}
\bibliography{scaling-looped}

% %%%%%%%%%%%%%%%%%%%%%%%%%%%%%%%%%%%%%%%%%%%%%%%%%%%%%%%%%%%%
\newpage
\appendix
\section{Appendix}\label{app:compute}
\paragraph{Compute.} The main scaling experiments were run on NVIDIA H100 GPUs (80GB). Approximate GPU-hours by experiment: isoFLOP scaling study ($\sim$1{,}200), $\mu$P transfer validation ($\sim$200), early-exit and loop-depth variants ($\sim$300), analysis ($\sim$100). Total: $\sim$2{,}000 GPU-hours. All training data fits within 2TB of storage. The multi-seed benchmark runs (Table~\ref{tab:bench_1e18}) and the 32-layer study (Appendix~\ref{app:32L}) were run on NVIDIA A100 GPUs (80GB).
\begin{figure}[H]
    \centering
    \includegraphics[width=0.5\linewidth]{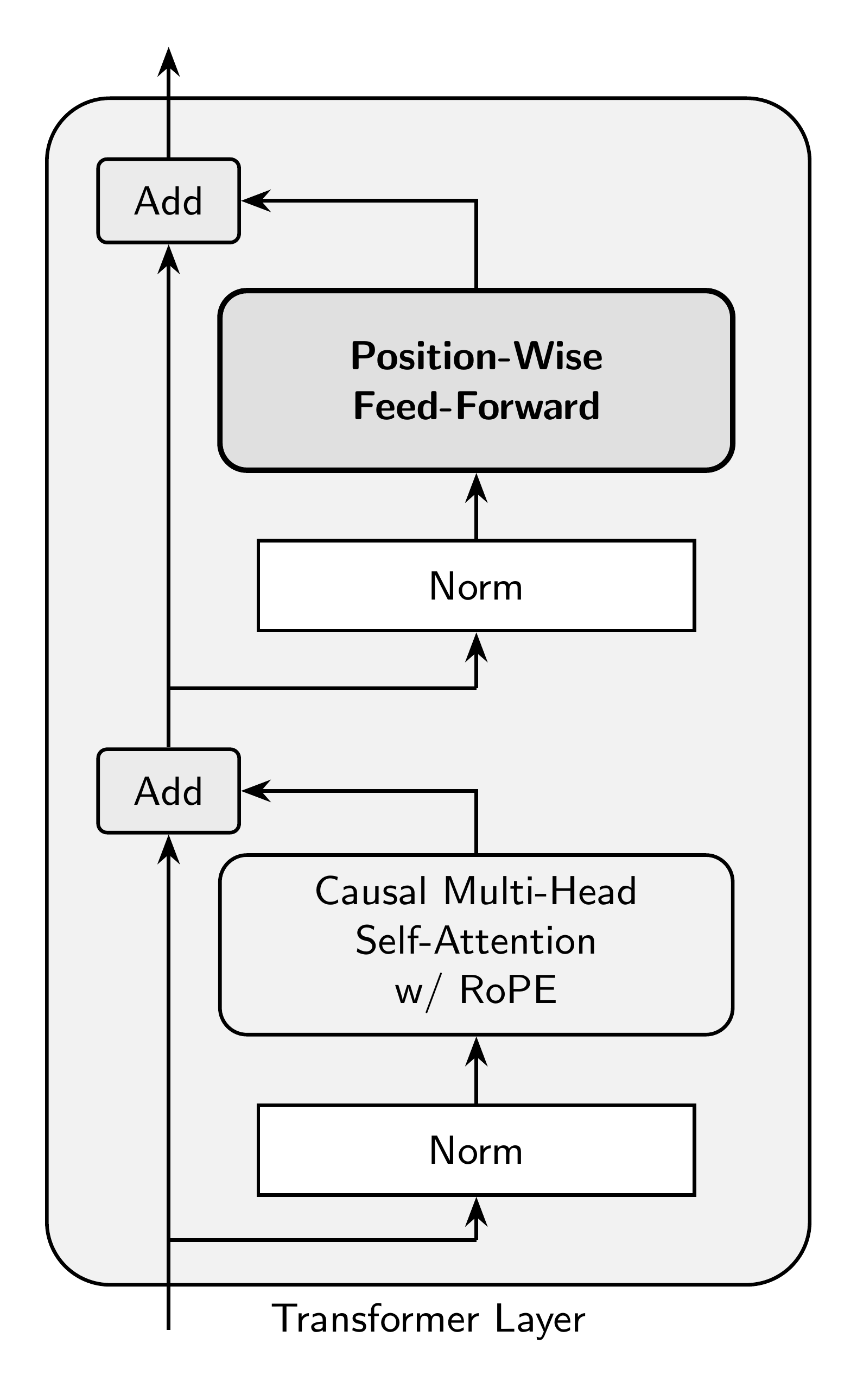}
    \caption{The pre-norm transformer layer, which we use in this study across all models. For our position-wise feed-forward network (FFN), we use SwiGLU. For the norms we use an RMS Norm without a learnable gain.}
    \label{fig:transformer_block}
\end{figure}

\begin{table}[H]
  \centering
  \caption{Model configurations in this study. $d_\text{ff}$ is rounded up to the nearest multiple of 64. Active parameters
  (including embeddings) are equal across all four architectures at each scale; stored parameters vary by architecture (see
  Table~\ref{tab:flops}). Some isoFLOP runs use intermediate widths between the sizes in this table. Parameter counts for MoE architectures are with $k{=}2$ active experts out of $E{=}8$ total.}
  \label{tab:model_configs}
  \begin{tabular}{rrrrrrr}
  \toprule
  $d_{\text{model}}$ & $d_{\text{ff}}$ & $n_{\text{heads}}$ & $w_{\text{ratio}}$ & Active (M) & Stored LMoE (M) & Stored MoE (M) \\
  \midrule
  128  & 384  & 2  & 1.0 & 16  & 18  & 23  \\
  256  & 704  & 4  & 2.0 & 39  & 45  & 65  \\
  384  & 1024 & 6  & 3.0 & 67  & 81  & 124 \\
  512  & 1408 & 8  & 4.0 & 103 & 129 & 207 \\
  640  & 1728 & 10 & 5.0 & 144 & 184 & 303 \\
  768  & 2048 & 12 & 6.0 & 190 & 247 & 417 \\
  896  & 2432 & 14 & 7.0 & 246 & 325 & 560 \\
  1024 & 2752 & 16 & 8.0 & 305 & 407 & 711 \\
  \bottomrule
  \end{tabular}
\end{table}

\paragraph{$\mu$P transfer at 16 layers.} We also spot-check transfer at the full 16-layer effective depth for Base and MoE with abbreviated pretraining runs (12{,}554 steps, WSD schedule): the base learning rate (0.01) is the argmin at $d_\text{model} \in \{128, 256\}$ and within 0.71\% (Base) and 0.06\% (MoE) of the best loss at $d_\text{model}=512$.

\begin{figure}[t]
    \centering
    \begin{subfigure}{0.48\linewidth}
        \centering
        \includegraphics[width=\linewidth]{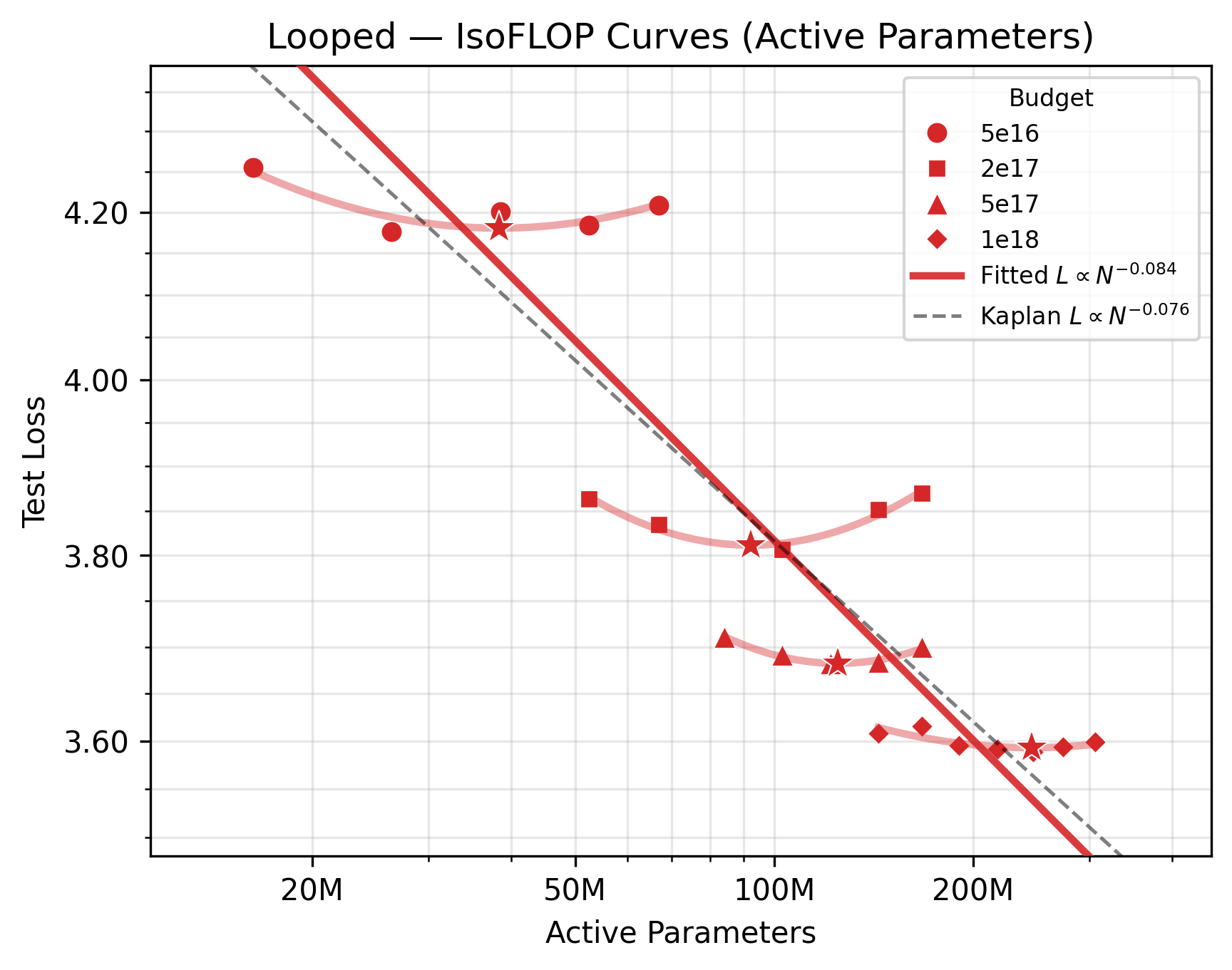}
        \caption{Looped}
    \end{subfigure}
    \hfill
    \begin{subfigure}{0.48\linewidth}
        \centering
        \includegraphics[width=\linewidth]{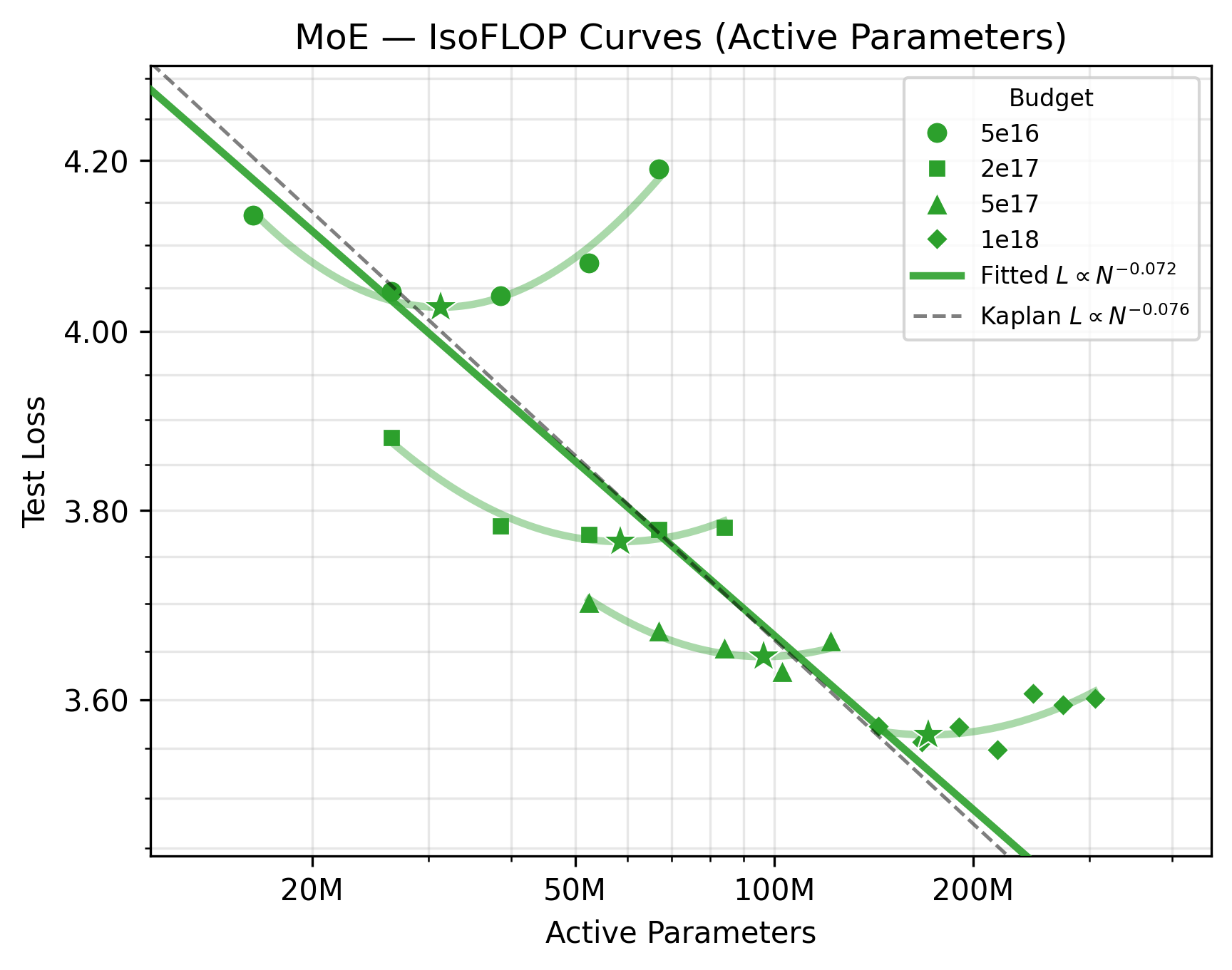}
        \caption{MoE}
    \end{subfigure}
    \caption{\textbf{Left:} IsoFLOP curves for Looped. \textbf{Right:} IsoFLOP curves for MoE. Power law fit to compute budgets from $5 \times 10^{16}$ to $10^{18}$ FLOPs. Solid lines show fitted power-law scaling relations; the dashed line shows the Kaplan et al. scaling exponent.}
    \label{fig:isoflops_individual_others}
\end{figure}

\begin{figure}[H]
    \centering
    \includegraphics[width=0.8\linewidth]{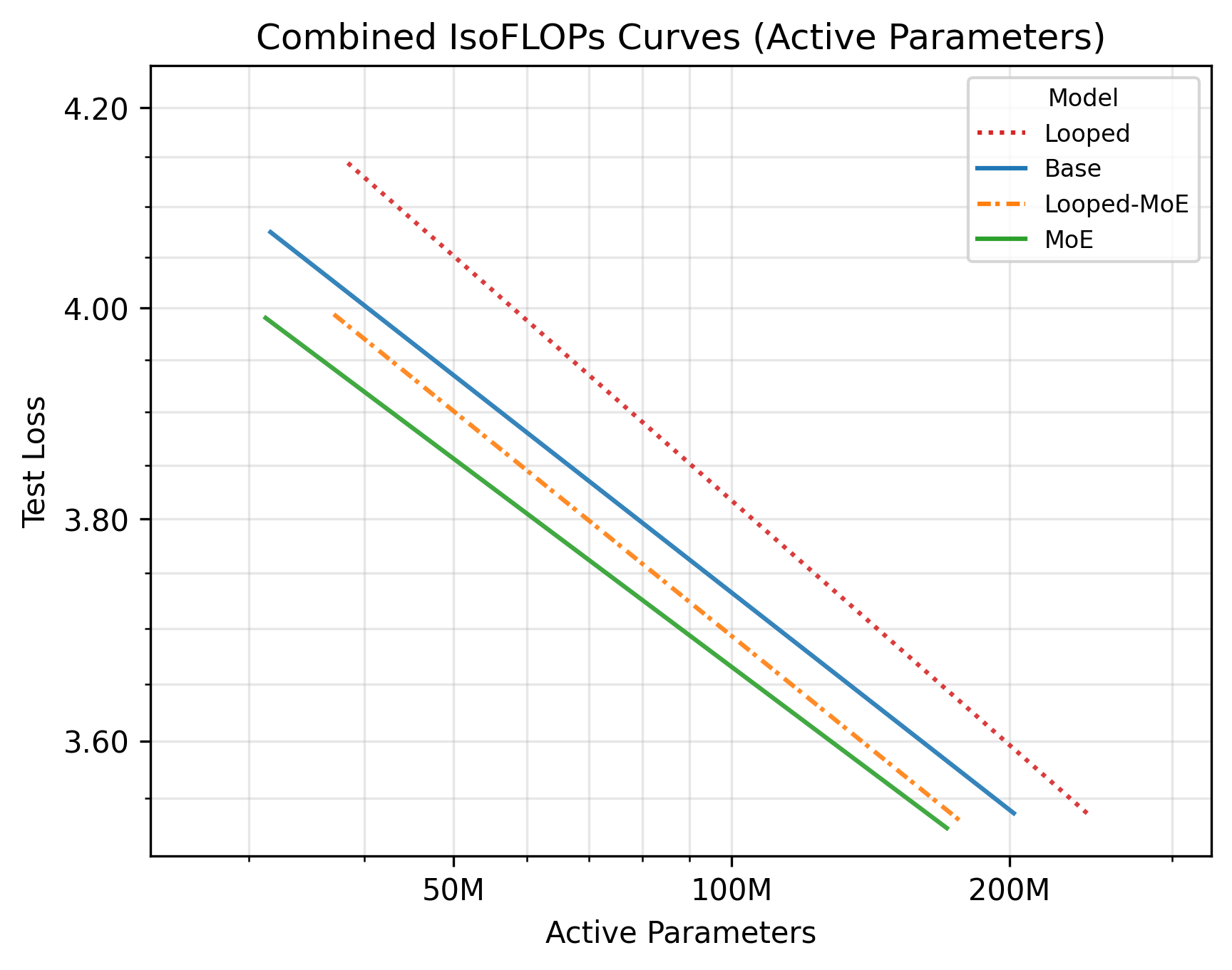}
    \caption{Combined scaling laws for all models in the study. Lower test loss is better. All architectures have roughly the same scaling slope, but are offset differently. MoE scales the best, followed by Looped-MoE, then Base and lastly Looped. Looping the same architecture seems thus to strictly reduce expressivity. However, in this paper we focus on how adding looping and MoE improves upon dense Base scaling and also Looped scaling.}
    \label{fig:isoflops_combined_all} 
\end{figure}
\newpage

\newcommand{\ms}[2]{\begin{tabular}[t]{@{}c@{}}#1\\[-2pt]{\scriptsize\color{gray}$\pm$#2}\end{tabular}}
\begin{table}[H]
  \centering
  \caption{Full AI2 OLMES benchmark results at $10^{18}$ FLOPs for all four architectures at their compute-optimal widths (mean $\pm$ std over 6 seeds). We report Core 8 (Core 9 without BoolQ) in the main text because BoolQ varies widely across seeds at this scale (e.g., MoE std of 10.3).}
  \label{tab:bench_1e18_full}
    \footnotesize\setlength{\tabcolsep}{2.2pt}
  \begin{tabular}{llccccccccc|cc}
    \toprule
    \textbf{Model} & \textbf{Params} & \textbf{ARC-E} & \textbf{ARC-C} & \textbf{BoolQ} & \textbf{CSQA} & \textbf{HellaSwag} & \textbf{OBQA} & \textbf{PIQA} & \textbf{SIQA} & \textbf{WinoG} & \textbf{Core 9} & \textbf{Core 8} \\
    \midrule
    Looped & 168M & \ms{41.1}{0.5} & \ms{24.4}{1.2} & \ms{59.4}{4.2} & \ms{\textbf{30.0}}{1.6} & \ms{25.5}{0.5} & \ms{26.5}{1.6} & \ms{56.6}{0.5} & \ms{\textbf{39.8}}{0.7} & \ms{51.2}{1.0} & \ms{39.4}{0.5} & \ms{36.9}{0.3} \\
    Base & 246M & \ms{\textbf{41.6}}{1.3} & \ms{24.7}{1.0} & \ms{60.8}{2.2} & \ms{28.5}{1.0} & \ms{26.3}{0.8} & \ms{27.1}{1.7} & \ms{\textbf{56.8}}{0.7} & \ms{39.5}{0.6} & \ms{51.2}{0.8} & \ms{39.6}{0.4} & \ms{36.9}{0.4} \\
    Looped-MoE & 216M & \ms{41.4}{0.9} & \ms{24.4}{0.4} & \ms{\textbf{63.4}}{0.7} & \ms{29.8}{1.2} & \ms{26.1}{0.4} & \ms{26.7}{1.4} & \ms{56.5}{0.6} & \ms{39.6}{0.8} & \ms{\textbf{52.3}}{0.9} & \ms{\textbf{40.0}}{0.3} & \ms{37.1}{0.4} \\
    MoE & 486M & \ms{41.5}{1.3} & \ms{\textbf{25.4}}{0.5} & \ms{54.2}{10.3} & \ms{29.5}{1.1} & \ms{\textbf{26.5}}{0.3} & \ms{\textbf{27.4}}{2.3} & \ms{56.6}{0.7} & \ms{39.6}{0.8} & \ms{51.4}{1.0} & \ms{39.1}{1.4} & \ms{\textbf{37.2}}{0.5} \\
    \bottomrule
  \end{tabular}
\end{table}

\begin{table}[H]
    \centering
    \caption{Perplexity at increasing levels of FLOPs saved via entropy-based early exit (training-free). The \emph{Full depth} column shows baseline perplexity with no early exit; each subsequent column shows perplexity when that fraction of FLOPs is skipped. Sparse experts alone do not improve early-exit tolerance --- MoE degrades faster than Base. The benefit is attributable specifically to looping: Looped achieves lower perplexity than both Base and MoE at every savings level, and the advantage grows with the number of loops. At 10\% FLOPs saved, Looped-MoE $2\times8$ reaches perplexity 42.0 versus 55.4 for Base and 75.7 for MoE.}
    \begin{tabular}{lccccc}
    \toprule
    Model & Full depth & 5\% & 10\% & 20\% & 30\% \\
    \cmidrule(lr){3-6}
    & & \multicolumn{4}{c}{FLOPs Saved} \\
    \midrule
    Base & 34.8 & 43.1 & 55.4 & 112.6 & 272.6 \\
    MoE & 34.8 & 51.9 & 75.7 & 167.5 & 369.2 \\
    Looped & 36.2 & 40.9 & 50.2 & 84.3 & 156.0 \\
    Looped-MoE & 35.9 & 41.1 & 51.0 & 89.7 & 182.6 \\
    Looped-MoE $4\times4$ & 35.4 & 38.2 & 44.3 & 71.0 & 160.1 \\
    Looped-MoE $2\times8$ & 37.3 & 38.8 & 42.0 & 57.6 & 153.5 \\
    \bottomrule
    \end{tabular}
    
    \label{tab:ee_perplexity}
\end{table}

\begin{table}[H]
    \centering
    \caption{Perplexity at increasing levels of FLOPs saved via confidence-based early exit (training-free): a token exits when the gap between its top-1 and top-2 output probabilities exceeds a threshold. Looped models exit at loop boundaries as in Table~\ref{tab:ee_perplexity}; Base and MoE exit at a single fixed point after layer 7 of 16, the best single exit layer for both models. Looped models degrade least, as with entropy.}
    \begin{tabular}{lccccc}
    \toprule
    Model & Full depth & 5\% & 10\% & 20\% & 30\% \\
    \cmidrule(lr){3-6}
    & & \multicolumn{4}{c}{FLOPs Saved} \\
    \midrule
    Base & 34.8 & 51.3 & 80.4 & 208.2 & 586.2 \\
    MoE & 34.8 & 51.0 & 76.3 & 178.7 & 440.9 \\
    Looped & 36.2 & 41.9 & 53.0 & 91.9 & 169.9 \\
    Looped-MoE & 35.9 & 41.9 & 54.1 & 99.2 & 196.7 \\
    Looped-MoE $4\times4$ & 35.4 & 39.2 & 48.6 & 89.2 & 199.9 \\
    Looped-MoE $2\times8$ & 37.3 & 40.0 & 47.2 & 79.3 & 176.6 \\
    \bottomrule
    \end{tabular}
    \label{tab:ee_confidence}
\end{table}

\newpage
\section{32-Layer Depth Study}\label{app:32L}
To test whether our findings hold beyond 16 effective layers, we repeat our procedure at 32 effective layers. For each architecture, we sweep the learning rate at the $\mu$P base width (0.005 is optimal for all), then run an isoFLOP sweep over widths at $10^{18}$ FLOPs and select the compute-optimal width. We compare Base and MoE (32 layers) with Looped and Looped-MoE ($16 \times 2$), plus Looped-MoE $8 \times 4$. Early exit uses the entropy criterion (Section~\ref{subsec:early-exit}). As at 16 effective layers, Looped-MoE achieves the highest Core 9 score (Table~\ref{tab:bench_32L}) and degrades least under early exit, with more looping ($8 \times 4$) further improving the trade-off (Table~\ref{tab:ee_32L}).

\paragraph{Number of experts.} We compare $E{=}8$ and $E{=}16$ at $k{=}2$ for Looped-MoE, holding active parameters and FLOPs matched ($k$ and expert size are unchanged; only the number of stored experts differs). Increasing $E$ from 8 to 16 mildly improves compute-optimal validation loss at both loop configurations ($16 \times 2$: 3.5698 to 3.5664; $8 \times 4$: 3.5723 to 3.5537).

\begin{table}[H]
  \centering
  \caption{32-layer AI2 OLMES benchmark results at each architecture's compute-optimal width, $10^{18}$ FLOPs.}
  \label{tab:bench_32L}
  \resizebox{\linewidth}{!}{%
  \begin{tabular}{lccccccccc|c}
    \toprule
    \textbf{Model} & \textbf{ARC-E} & \textbf{ARC-C} & \textbf{BoolQ} & \textbf{CSQA} & \textbf{HellaSwag} & \textbf{OBQA} & \textbf{PIQA} & \textbf{SIQA} & \textbf{WinoG} & \textbf{Core 9} \\
    \midrule
    Looped $16 \times 2$ d640 & 40.7 & 25.1 & \textbf{63.7} & 29.3 & \textbf{25.9} & \textbf{27.6} & \textbf{56.8} & 39.3 & 49.3 & 39.74 \\
    Base d512 & 42.0 & 25.2 & 49.5 & \textbf{30.1} & 25.4 & 27.2 & 56.5 & \textbf{40.1} & \textbf{52.5} & 38.71 \\
    Looped-MoE $16 \times 2$ d512 & \textbf{42.5} & \textbf{26.6} & 63.5 & 29.9 & 25.6 & 25.2 & 56.1 & 38.8 & 51.6 & \textbf{39.98} \\
    MoE d512 & 40.7 & 23.6 & \textbf{63.7} & 28.9 & 25.7 & \textbf{27.6} & 54.5 & 39.4 & 50.7 & 39.43 \\
    \bottomrule
  \end{tabular}%
  }
\end{table}

\begin{table}[H]
  \centering
  \caption{Early exit at 32 effective layers (entropy criterion). Perplexity at increasing compute saved, relative to each model's full forward pass. E16k2 denotes $E{=}16$ experts with $k{=}2$ active; all other MoE models use $E{=}8$, $k{=}2$.}
  \label{tab:ee_32L}
  \begin{tabular}{lcccc}
    \toprule
    Model & Full depth & 10\% & 20\% & 30\% \\
    \midrule
    Base d512 & 34.75 & 49.67 & 88.42 & 183.67 \\
    Looped-MoE $16 \times 2$ d512 & 35.51 & 47.82 & 78.31 & 144.63 \\
    Looped-MoE $8 \times 4$ E16k2 d640 & 34.94 & \textbf{39.76} & \textbf{51.75} & \textbf{83.61} \\
    MoE d512 & 35.32 & 54.41 & 86.50 & 159.21 \\
    \bottomrule
  \end{tabular}
\end{table}

\newpage
\end{document}